\documentclass[10pt,twocolumn,letterpaper]{article}
\usepackage[pagenumbers]{cvpr}

\usepackage{graphicx}
\usepackage{amsmath}
\usepackage{amssymb}
\usepackage{booktabs}
\usepackage{xcolor}
\usepackage{multirow}
\usepackage{hyperref}

\AtBeginDocument{%
  }

\title{The Aging Multiverse: Generating Condition-Aware Facial Aging Tree via Training-Free Diffusion}

\author{
Bang Gong$^{1}$\footnotemark[1] \quad 
Luchao Qi$^{1}$\footnotemark[1] \quad
Jiaye Wu$^{2}$ \quad 
Zhicheng Fu$^{3}$ \\
Chunbo Song$^{3}$ \quad 
David W. Jacobs$^{2}$ \quad 
John Nicholson$^{3}$ \quad 
Roni Sengupta$^{1}$ \\
$^{1}$UNC Chapel Hill \quad 
$^{2}$University of Maryland \quad 
$^{3}$Lenovo \\
{\tt\small gongbang@cs.unc.edu, lqi@cs.unc.edu, jiayewu@cs.umd.edu} \\
{\tt\small zcfu@motorola.com, csong2@lenovo.com, dwj@umd.edu, jnichol@lenovo.com, ronisen@cs.unc.edu}
}

\usepackage[capitalize]{cleveref}

\usepackage[colorinlistoftodos,prependcaption,textsize=tiny,draft]{todonotes}

\crefformat{section}{\S#2#1#3} 
\crefformat{subsection}{\S#2#1#3}
\crefformat{subsubsection}{\S#2#1#3}

\usepackage{multirow}
\usepackage{makecell}
\usepackage{enumitem}
\usepackage[percent]{overpic}
\usepackage{lipsum}
\usepackage[dvipsnames]{xcolor}
\usepackage{annotate-equations}
\usepackage{caption}
\usepackage{subcaption}
\usepackage{graphicx}

\let\titleold\title
\renewcommand{\title}[1]{\titleold{#1}\newcommand{\thetitle}{#1}}

\usepackage{colortbl}
\definecolor{tabfirst}{rgb}{1, 0.7, 0.7} 
\definecolor{tabsecond}{rgb}{1, 0.85, 0.7} 
\definecolor{tabthird}{rgb}{1, 1, 0.7} 

\begin{document}

\twocolumn[{%
\renewcommand{\twocolumn}[1][]{#1}%
\maketitle
\vspace{-2.5em}
\includegraphics[width=0.45\linewidth]{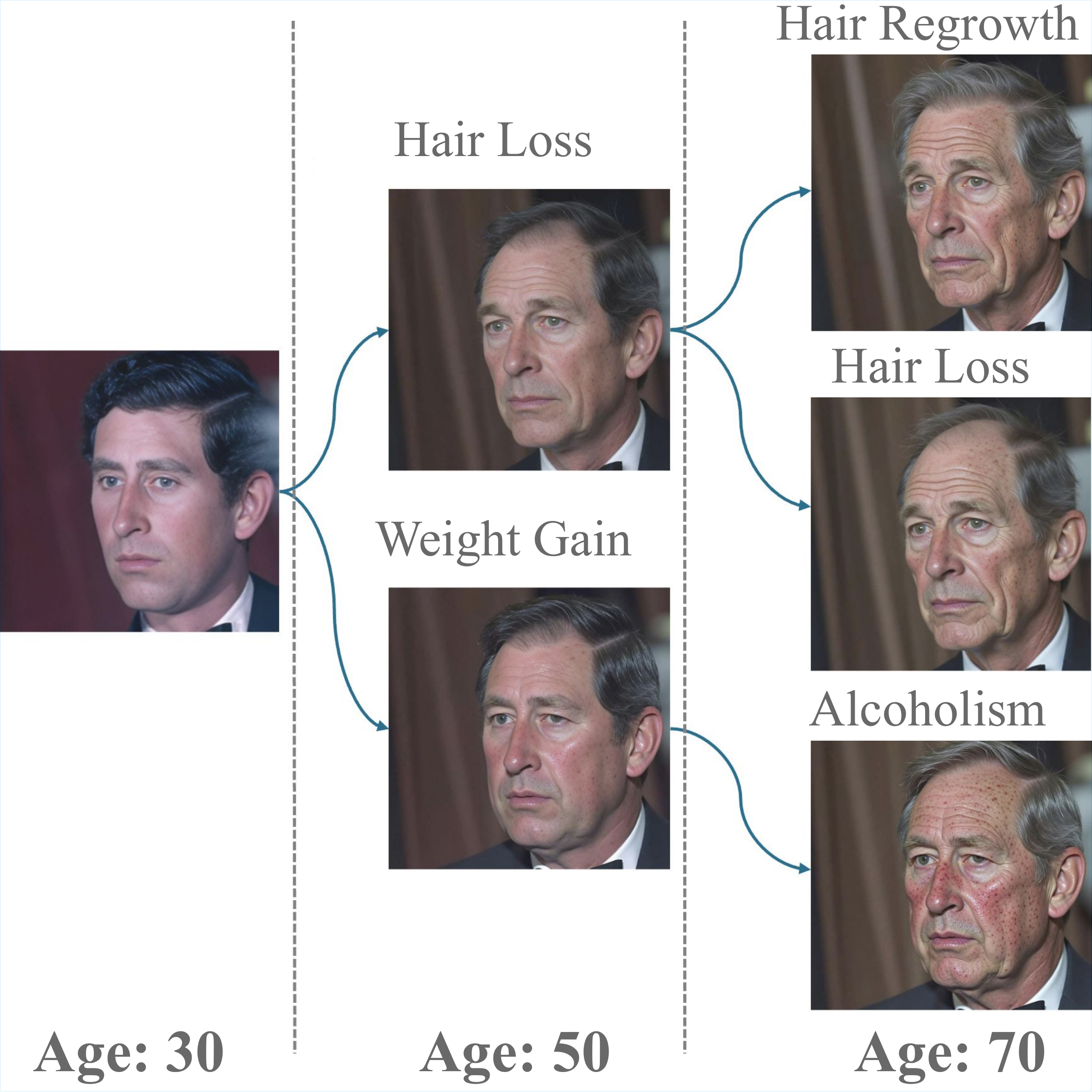} \hspace{2em}
\raisebox{0.3ex}{\includegraphics[width=0.45\linewidth]{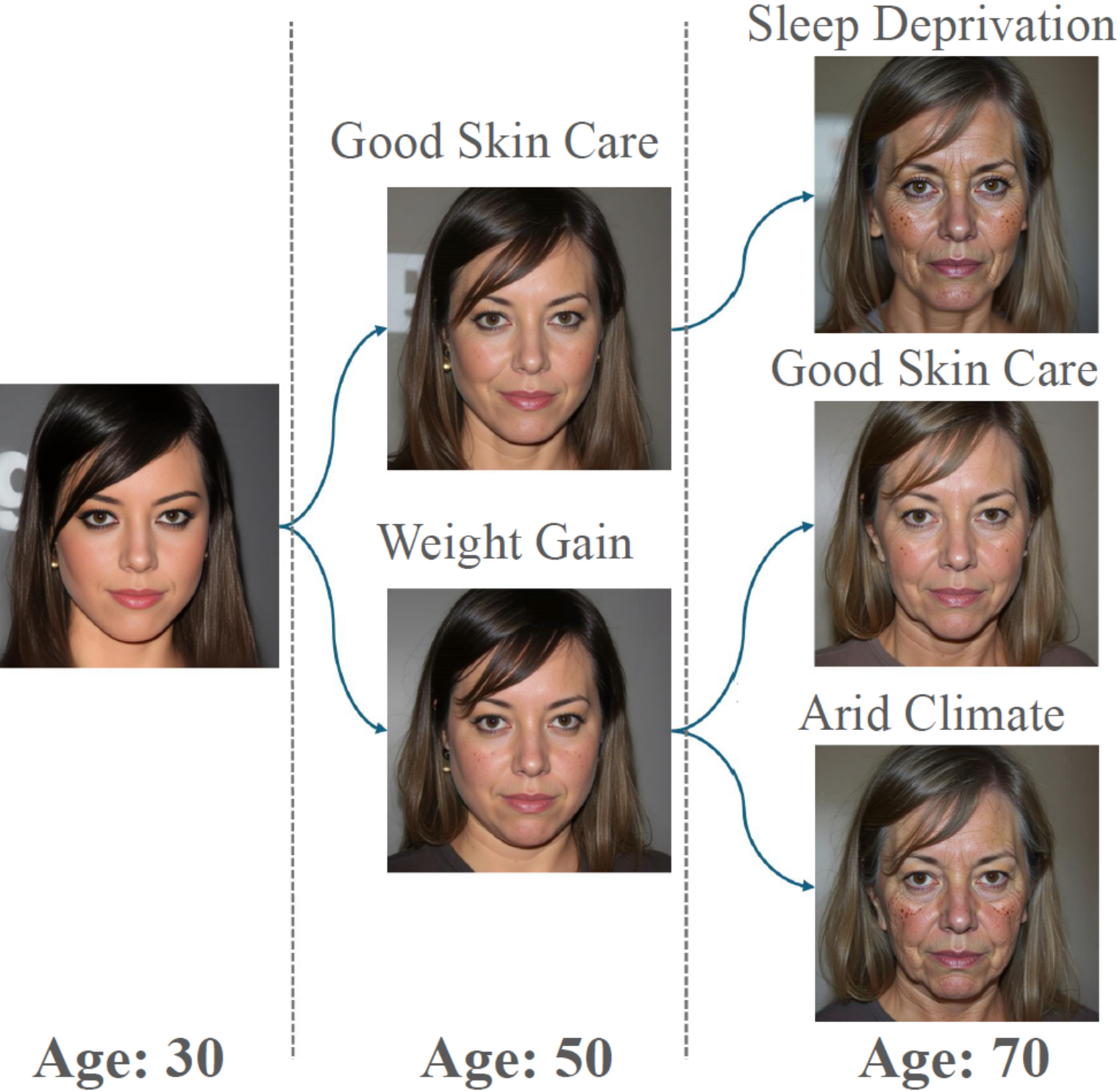}}
\captionof{figure}{
Given a single input image, our method generates an aging multiverse—multiple plausible aging trajectories conditioned on different external factors such as weight gain, skincare, hair loss, alcohol use, and environment.
Each branch visualizes how environment, lifestyle, and health choices could shape appearance over time.
}
\label{fig:teaser}
}]

\footnotetext[1]{* Equal contribution.}

\begin{abstract}
  We introduce the Aging Multiverse, a framework for generating multiple plausible facial aging trajectories from a single image, each conditioned on external factors such as environment, health, and lifestyle. Unlike prior methods that model aging as a single deterministic path, our approach creates an aging tree that visualizes diverse futures.
To enable this, we propose a training-free diffusion-based method that balances identity preservation, age accuracy, and condition control. Our key contributions include attention mixing to modulate editing strength and a Simulated Aging Regularization strategy to stabilize edits. Extensive experiments and user studies demonstrate state-of-the-art performance across identity preservation, aging realism, and conditional alignment, outperforming existing editing and age-progression models, which often fail to account for one or more of the editing criteria. By transforming aging into a multi-dimensional, controllable, and interpretable process, our approach opens up new creative and practical avenues in digital storytelling, health education, and personalized visualization. Additional visual examples are available on our project website: \textcolor{blue}{\href{https://agingmultiverse.github.io/}{https://agingmultiverse.github.io/}}
\end{abstract}

\vspace{-1em}
\section{Introduction}
\label{sec:intro}
What might you look like in your 60s? Would a consistent skincare routine make you appear more youthful? How would hair loss or alcohol addiction affect your appearance over time? While genetics plays a significant role in facial aging, external factors such as environmental exposure (e.g., sunlight, humidity), health conditions (e.g., stress, alcoholism, weight changes), and daily habits (e.g., skincare) can profoundly influence how we age.
In this paper, we introduce the concept of an aging multiverse—a framework for generating multiple plausible facial aging trajectories for an individual, each conditioned on different external factors. This approach enables a range of creative and practical applications. By transforming aging into a multi-dimensional, controllable, and interpretable process, the aging multiverse allows users to explore an “aging tree” of lifestyle-driven futures, empowering applications in digital storytelling, health education, and personalized visualization.

Prior portrait image age transformation methods~\cite{alaluf_only_2021, avidan_agetransgan_2022, Chen_2023_BMVC, wahid2024diffage3ddiffusionbased3dawareface, gomez-trenado_custom_2022} mostly focus on learning a global aging prior through pretraining on large human face datasets like FFHQ~\cite{karras_style-based_2019}. These approaches often do not consider the inherent plurality of the aging process, and when they do~\cite{li_pluralistic_2023}, they do not condition it on physical external factors that affect aging.  In contrast to these approaches that generate only ``aging line", we aim to generate an ``aging tree" by developing a novel training-free conditional method.

Generating condition-aware aging paths from an input image requires strong preservation of identity, while editing aging features and introducing the specific attributes aligned with the conditions.  While existing face transformation techniques~\cite{Chen_2023_BMVC, avidan_agetransgan_2022, alaluf_only_2021} have now excelled in identity-preserving aging, they are unable to add any condition to their generation. Existing image-editing approaches~\cite{rout_semantic_2024, deng2024fireflowfastinversionrectified, wang_taming_2024}, on the other hand, can handle different external conditions but struggle to balance all three criteria, e.g., RF-Solver-Edit~\cite{wang_taming_2024} can preserve identity but struggles in aging and alignment to condition, and FlowEdit~\cite{kulikov_flowedit_2024} can enable condition-aware aging but struggles to preserve identity.
Our goal of generating an aging multiverse requires jointly editing age and conditional attributes while maintaining identity, evolving the conventional inversion-editability trade-off into a three-way balance among identity, age, and external conditions. To solve this problem, we propose a novel training-free method that can transform an input image into any target age under any condition defined in a text prompt. 

Our key technical contribution lies in introducing training-free attention mixing and regularization strategies to enable multi-factorial aging while balancing identity preservation, age accuracy, and condition control. Central to our approach is the observation that the alignment between attention features for identity inversion and those for condition-aware editing determines the editability-identity trade-off. We leverage this by amplifying editing signals when these features align and attenuating them when they conflict. Building on this insight, we propose two modulation functions that operate on the Value and Key tensors of the attention blocks in a second-order Rectified Flow model~\cite{wang_taming_2024}. Additionally, we introduce Simulated Aging Regularization, which applies unconditioned age progression to derive a stable aging trajectory, serving as a guide to further regularize attention features during condition-aware editing.

Our technique is training-free, and can be applied to any individual to simulate any age between 20-90 years old with external factors related to the environment, health, and lifestyle of the individual. This is in contrast to existing face age transformation techniques that either rely on global face aging datasets ~\cite{Chen_2023_BMVC} or personalized datasets for training \cite{qi_mytimemachine_2024}. 
This training-free property ensures a plug-and-play framework, enabling universal compatibility with existing DiT-based models~\cite{esser_scaling_2024} and reduced computational resources.

We conduct comprehensive evaluations on both celebrity and non-celebrity images, using a combination of automated metrics and user studies to assess identity preservation, age accuracy, and consistency with the specified external conditions. Our evaluation indicates that our approach can provide the best balance across all three axes of image editing, while previous methods mostly succeed on one or two of them and fail for the rest. We further ablate the importance of our proposed attention mixing of key and value tensors, as well as the effectiveness of our attention regularization via simulated unconditional aging.


In summary, our key contributions are: (i) We formulate novel problem of generating an aging multiverse—or an aging tree—for an individual from a single image, simulating appearances across a wide age range (20–90) under varying environmental, health, and lifestyle conditions, a first to our knowledge. (ii) We propose a novel training-free framework that balances three key objectives: identity preservation, aging accuracy, and adherence to external conditions. (iii) We introduce attention mixing and attention regularization strategies that significantly improve the inversion-editability trade-off, leading to state-of-the-art performance on both celebrity and non-celebrity images.

\begin{figure*}[!ht]
    \centering    \includegraphics[width=\linewidth]{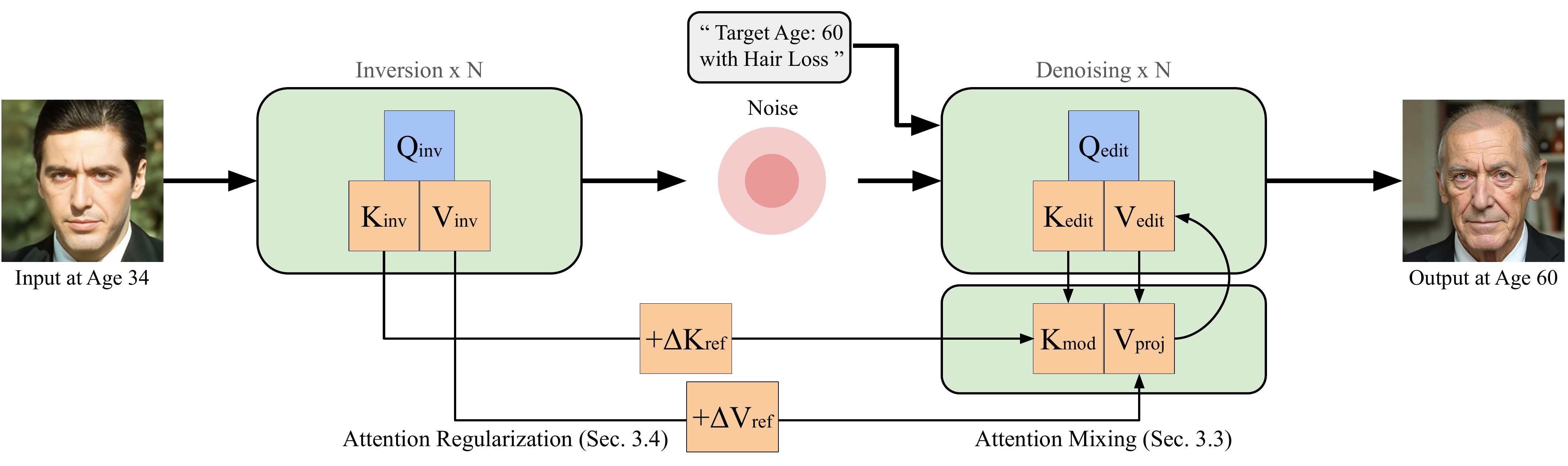}
    \vspace{-1.5em}
    \caption{
    Overview of our training-free conditional-age progression framework. Given an input image and a textual description of external aging factors, our method leverages flow matching techniques to perform editing. Our approach balances three competing objectives—identity preservation, age accuracy, and condition alignment—enabling conditional age transformation without retraining.
    Our key innovations are: (i) attention mixing (\cref{ssec:att-mix}) of Key and Value tensors between inversion and editing, and (ii) attention regularization (\cref{ssec:att-mix}) with simulated unconditional aging to achieve the best inversion-editability trade-off.
    }
    \label{fig:method}
    \vspace{-1.5em}
\end{figure*}

\section{Related Work}
\label{sec:related_work}

\noindent
\textbf{Age Transformation.}
Earlier approaches for face age transformation relied heavily on Generative Adversarial Networks (GANs), particularly StyleGAN2~\cite{karras_analyzing_2020}, due to its disentangled latent space, leading to techniques that perform linear age-editing \cite{shen_interfacegan_2020,nitzan_large_2022} and later non-linear transformations~\cite{alaluf_only_2021,gomez-trenado_custom_2022,avidan_agetransgan_2022,qi_mytimemachine_2024}. Recent methods have explored the Stable Diffusion model for age editing using age as a prompt in DiffAge3D~\cite{wahid2024diffage3ddiffusionbased3dawareface} and FADING~\cite{Chen_2023_BMVC}. However, these approaches lack the ability to generate an ``aging tree" by performing condition-aware aging. While PADA~\cite{li_pluralistic_2023} models the stochastic nature of aging by introducing diversity in the diffusion latent space, it does not support conditional guidance like specific lifestyle or health attributes. In contrast, our method enables diverse, controllable age transformations conditioned on both target age and external factors, offering a more flexible and open framework for multifaceted facial aging.

\vspace{1mm}
\noindent
\textbf{The Identity–Editability Trade-off in Diffusion Models.}
Diffusion-based editing method often first inverts the input image into a noisy latent and then denoises it with text prompt guidance. A key challenge lies in achieving faithful modifications to a specific attribute (e.g., age) while preserving the identity and structure of the original image. Early methods like DDIM inversion~\cite{song_denoising_2022} and SDEdit~\cite{meng_sdedit_2022} enabled diffusion-based editing but struggled with identity drift due to error accumulation. Subsequent techniques such as NTI~\cite{mokady_null-text_2022}, Negative-Prompt Inversion~\cite{miyake2024negativepromptinversionfastimage}, and ReNoise~\cite{garibi2024renoise} improved inversion fidelity and stability. To better balance identity and editability, recent works proposed manipulating internal representations during denoising—Prompt-to-Prompt~\cite{hertz_prompt--prompt_2022}, Plug-and-Play~\cite{tumanyan_plug_and_play}, and MasaCtrl~\cite{cao_2023_masactrl} used attention control, while Asyrp~\cite{kwon_diffusion_2023} and Pix2Pix-Zero~\cite{parmar_zero-shot_2023} introduced latent and semantic guidance strategies.




Flow-based diffusion models like Rectified Flow~\cite{lipman_flow_2023, liu_flow_2022} have enabled faster and more stable editing, with works such as RF-Inversion~\cite{rout_semantic_2024}, FireFlow~\cite{deng2024fireflowfastinversionrectified}, and FTEdit~\cite{xu2024unveil} enhancing inversion quality through novel solvers, dynamic control, and fixed-point refinements. To improve the identity-editability trade-off, methods like RF-Solver-Edit~\cite{wang_taming_2024}, FireFlow~\cite{deng2024fireflowfastinversionrectified}, and KV-Edit~\cite{zhu2025kv} introduced feature sharing strategies, including attention reuse and memory-efficient KV caching inspired by language models.

Complementary works have also addressed architectural limits on editing fidelity: Stable Flow~\cite{avrahami_stable_2024} and HeadRouter~\cite{xu2024headrouter} proposed routing and reweighting based on attention bottlenecks, while Fluxpace~\cite{dalva_fluxspace_2024} and FlowChef~\cite{patel2024flowchef} enabled controllable edits via latent direction extraction and optimization techniques.



While prior methods improve either fidelity or controllability, few consider age editing as a core task, and even fewer address the joint challenge of preserving identity, age accuracy, and external conditions. Our work addresses this gap with a novel training-free framework that introduces attention mixing and regularization techniques for generating condition-aware aging trajectories.

\vspace{-0.5em}
\section{Method}
\label{sec:method}

We first introduce some preliminaries about Rectified Flow in Sec. \ref{ssec:prelim}. We then introduce key technical innovations of our pipeline in the following sections, starting with auto-generating a detailed prompt to create condition-aware aging using LLM in Sec. \ref{ssec:prompt}, followed by attention mixing for inversion-editability trade-off in Sec. \ref{ssec:att-mix}, and simulated aging regularization for improving editng stability and robustness in Sec. \ref{ssec:sar}.

\subsection{Preliminaries}
\label{ssec:prelim}
Rectified Flow (RF) models aim to learn a mapping between the real data distribution $\pi_0$ and a Gaussian noise distribution $\pi_1$ by modeling a velocity field $v$. This mapping is formulated as an ordinary differential equation (ODE):
\begin{align}
\frac{dZ_t}{dt} = v(Z_t, t), \quad t \in [0, 1], \label{eq:rf_ode}
\end{align}
where $v(Z_t, t)$ denotes the velocity field at time $t$ and state $Z_t$. In practice, this field is parameterized using DiT architectures~\cite{flux2024, esser_scaling_2024}. By design, the intermediate state $Z_t$ follows a linear interpolation between $X_0 \sim \pi_0$ and $X_1 \sim \pi_1$. Formally:
\begin{align}
Z_t \sim (1-t)X_0 + tX_1. \label{eq:rf_interpolate}
\end{align}

During sampling, the process begins with $Z_1 \sim \mathcal{N}(0, I)$. Given discretization steps $\{t_N, t_{N-1}, \ldots, t_0\}$, the ODE in Eq.~\eqref{eq:rf_ode} is solved numerically as:
\begin{align}
Z_{t_{i-1}} = Z_{t_i} + \int_{t_i}^{t_{i-1}} v_\theta(Z_\tau, \tau)\, d\tau, \quad i = N, N-1, \ldots, 1, \label{eq:rf_integral}
\end{align}
where $v_\theta$ is the learned velocity field.

To enable image editing with Rectified Flow models, prior work~\cite{rout_semantic_2024, wang_taming_2024} typically follows two steps:  
1) \textbf{Inversion} maps the input image to the noise space using a diffusion transformer  $\text{DiT}$ as $Z_t = \text{DiT}(Z_{t-1}, t)$.
2) \textbf{Editing} performs denoising conditioned on a target text prompt ($\textit{txt}$) and the inverted noise as $Z_{t-1} = \text{DiT}(Z_t,\ t,\ \text{txt})$. During inversion, the diffusion transformer consists of multiple attention blocks with the query, key, and value as $(Q_{\text{inv}},\ K_{\text{inv}},\ V_{\text{inv}})$. Similarly, during editing, the diffusion transformer has multiple attention blocks that depend on the text prompt ($\textit{txt}$) with the query, key, and value as  $(Q_{\text{edit}},\ K_{\text{edit}},\ V_{\text{edit}})$. To achieve editing that preserves the identity of the original image, RF-Solver-Edit~\cite{wang_taming_2024} proposed replacing the value of editing attention layers with that of the inversion attention layer as:
\begin{align}
    (Q_{\text{edit}},\ K_{\text{edit}},\ V_{\text{edit}}) &\leftarrow (Q_{\text{edit}},\ K_{\text{edit}},\ V_{\text{inv}}). \label{eq:attn_replace}
\end{align}

While this replacement improves identity and background fidelity, it often results in overfitting to the original image $X^0$, particularly in the facial region. As a consequence, the intended semantic edits are suppressed, as illustrated in Fig.~\ref{fig:rf_comparison}, where RF-Solver-Edit~\cite{wang_taming_2024} produces minimal visible changes to the input.
Our approach also uses a similar Rectified Flow model, but proposes novel training-free attention feature modulation techniques that can balance identity preservation, age editing, and external conditions.

\subsection{Prompt Refinement}
\label{ssec:prompt}

While modern text-to-image (T2I) models generate highly detailed outputs, they often struggle to interpret complex or abstract conditions~\cite{kang2025flux}. For instance, prompting Flux with "\textit{a photo of a male at 40 years old addicted to alcohol}" fails to produce the expected facial characteristics. This is because high-level conditions like "alcohol addiction" are not directly grounded in visual features without additional context.

To bridge this gap, we refine prompts using GPT-4o~\cite{openai2024gpt4ocard}, which helps translate abstract conditions into specific, low-level facial attributes. Following prior work~\cite{fu2024mgie}, we use LLMs to expand vague inputs into detailed descriptions that guide the model more effectively. For example, we convert the original prompt into a refined version like "\textit{a 40-year-old man with pale skin, sunken eyes, and facial wrinkles due to long-term alcohol abuse}," enabling the model to better align appearance with the intended condition.

\begin{table*}[t]
\centering
\resizebox{0.8\textwidth}{!}{%
        \begin{tabular}{lcccccc}
            \toprule
            Method &
            \multicolumn{2}{c}{External Condition} &
            \multicolumn{2}{c}{Age Accuracy} &
            \multicolumn{2}{c}{ID preservation} \\
            \cmidrule(r){2-3} \cmidrule(r){4-5} \cmidrule(r){6-7}
            & CLIP-T($\uparrow$) & Human Eval.($\uparrow$) 
            & Age$_{MAE}$($\downarrow$) & Human Eval.($\uparrow$)  
            & ID$_{sim}$($\uparrow$) & Human Eval.($\uparrow$)  \\
            \midrule
            FADING* & - & - & \cellcolor{tabfirst}8.6 & \cellcolor{tabsecond}3.81 & \cellcolor{tabfirst}0.57 & \cellcolor{tabthird}3.82\\
            \midrule
            RF-Inversion & 0.299 & - & 13.9 & - & 0.34 & -\\
            FlowChef & 0.293 & - & 14.1 & - & 0.43 & - \\
            Fireflow & 0.299 & - & 16.5 & - & \cellcolor{tabsecond}0.51 & -\\
            FlowEdit & \cellcolor{tabthird}0.303 & \cellcolor{tabthird}3.30 & 13.4 & \cellcolor{tabthird}3.72 & 0.42 & 3.28\\
            RF-Solver-Edit & 0.292 & 3.16 & 17.8 & 3.08 & \cellcolor{tabfirst}0.57 & \cellcolor{tabfirst}3.89\\
            \midrule
            Ours w/o SAR & \cellcolor{tabsecond}0.322 & \cellcolor{tabsecond}3.57 & \cellcolor{tabthird}11.0 & 3.63 & 0.48 & 3.48\\
            Ours & \cellcolor{tabfirst}0.326 & \cellcolor{tabfirst}3.65 & \cellcolor{tabsecond}9.5 & \cellcolor{tabfirst}3.84 & \cellcolor{tabthird}0.49 & \cellcolor{tabsecond}3.84\\
            \bottomrule
        \end{tabular}%
        }
        \caption{
Quantitative comparison of condition-aware age transformation on celebrity data. 
An asterisk (\textbf{*}) indicates methods that require aging-specific pre-training and cannot be conditioned on external prompts; thus, no CLIP-T score is reported for these methods.
\colorbox{tabfirst}{\raisebox{0pt}[1ex][0ex]{Red}} highlights the best result, 
\colorbox{tabsecond}{\raisebox{0pt}[1ex][0ex]{Orange}}  indicates the second best, 
and \colorbox{tabthird}{\raisebox{0pt}[1ex][0ex]{Yellow}} denotes the third best.
        }
        \label{tab:celeb}
\end{table*}

\subsection{Attention Mixing For Inversion-Editability Trade-off}
\label{ssec:att-mix}

\begin{figure}[!t]
    \centering
    \vspace{-1.5em}
    \includegraphics[width=\columnwidth]{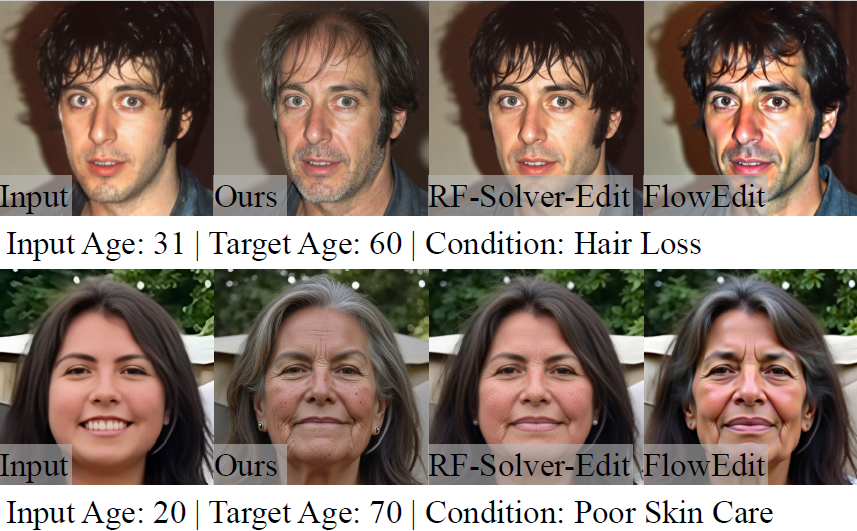}
    \vspace{-2em}
    \caption{Visual comparison of our method with RF-Solver-Edit~\cite{wang_taming_2024} and FlowEdit~\cite{kulikov_flowedit_2024}. Given input images, we edit the faces with a desired aging condition. RF-Solver-Edit shows limited editing capability, yielding results very similar to the input. FlowEdit can generate some edits but leads to id drift and unrealistic skin texture. In contrast, our method achieves stronger edits that accurately reflect the prompt.}
    \label{fig:rf_comparison}
    \vspace{-1em}
\end{figure}

Our goal is to enable text-driven age transformation while preserving input identity. A naive baseline is RF-Solver-Edit, which replaces self-attention values during denoising with those saved during inversion. This preserves identity and background well, but suppresses edits, since $V_{\text{edit}}$—which encodes the desired transformation—is fully discarded. As shown on the top of Fig.~\ref{fig:rf_comparison}, editing an image of \textit{Al Pacino} to appear ``60 years old with hair loss'' yields minimal visual change: identity is intact, but the edit fails.

\noindent \textit{Attention Value Projection.} To address the above limitation, we propose a modulation-based fusion of $V_{\text{inv}}$ and $V_{\text{edit}}$ that allows us to retain identity while enabling stronger edits, balancing the inversion-editing trade-off. The core idea is intuitive: if the edit direction aligns with the identity features, we amplify it; if not, we suppress it to avoid identity loss. We formalize this by computing the orthogonal vector projection of $V_{\text{inv}}$ onto $V_{\text{edit}}$ as:

\begin{equation}
V_{\text{proj}} = \alpha V_{\text{edit}}, \quad \text{where } \alpha = \frac{\langle V_{\text{inv}}, V_{\text{edit}} \rangle}{\langle V_{\text{edit}}, V_{\text{edit}} \rangle}
\label{eq:feature_projection}
\end{equation}

\noindent \textit{Text Embedding Masking.} Since DiT blocks jointly process text and image embeddings, we further refine our projection strategy to preserve guidance from the text prompt. Specifically, we mask out the text channels of $V_{\text{inv}}$ when computing $\alpha$ in Eq.~\ref{eq:feature_projection}, ensuring that the projection only fuses image-related features. After computing $\alpha$, we restore the text channels by setting their corresponding values in $\alpha$ to 1, thereby preserving prompt semantics from $V_{\text{edit}}$ in $V_{\text{proj}}$.





\noindent \textit{Attention Key Modulation.} Modifying only the attention value $V$ can cause inconsistencies. In particular, since $Q_{\text{edit}}$ and $K_{\text{edit}}$ do not carry information from the inversion branch, the attention weights may misalign with the modulated $V_{\text{proj}}$, leading to unrealistic results such as distorted face or out-of-distribution images, as shown in ablation studies Fig.~\ref{fig:k_motivation}. To address this, we also adjust the key tensor $K_{\text{edit}}$ using features from $K_{\text{inv}}$:

\begin{equation}
K_{\text{mod}} = K_{\text{edit}} + g \cdot (A \cdot K_{\text{inv}}), \quad \text{where } A = \text{softmax}\left( \frac{K_{\text{edit}} K_{\text{inv}}^T}{\sqrt{d_K}} \right)
\end{equation}

Here, $g$ is a scaling factor that controls the inversion-editing trade-off and $A$ serves as an attention alignment matrix. We used $g=0.5$ in this paper, which we empirically found to strike a good balance between editing and identity preservation. This update ensures that the attention computation reflects both the editing goal and the identity constraint, improving consistency in the final output.
Finally, attention layers during editing using $\text{DiT}$ is computed using $(Q_{\text{edit}}, K_{\text{mod}}, V_{\text{proj}})$.

\subsection{Simulated Aging Regularization in Attention Layers}

\label{ssec:sar}    
Our attention mixing technique enhances editing by enabling fine-grained control over attention distributions. However, directly modifying attention can lead to unstable results or out-of-distribution generations~\cite{dombrowski2024trade, yan2024diffusion}. To address this, we introduce a Simulated Aging Regularization mechanism that leverages reference-based "age directions", derived by simulating age clusters, to guide attention feature modifications, resulting in a semantically grounded and stable editing.

Specifically, we generate unconditional age-progressed images from the input using GPT-4o~\cite{hurst2024gpt}, and diversify them with Arc2Face~\cite{papantoniou_arc2face_2024} to construct distinct clusters corresponding to both the input and target age ranges. From these clusters, we compute average self-attention features for representative older (e.g., age 70) and younger (e.g., age 30) groups, denoted as $\overline{V}_{70\, \text{cluster}}$, $\overline{K}_{70\, \text{cluster}}$, $\overline{V}_{30\, \text{cluster}}$ and $\overline{K}_{30\, \text{cluster}}$ respectively. We then define a semantic aging direction in the attention space as:



\begin{align}
\Delta V_{\text{ref}} &= \overline{V}_{70\, \text{cluster}} - \overline{V}_{30\, \text{cluster}} \label{eq:delta_v} \\
\Delta K_{\text{ref}} &= \overline{K}_{70\, \text{cluster}} - \overline{K}_{30\, \text{cluster}} \label{eq:delta_k}
\end{align}

Given the age of the input image \( Age_{\text{input}} \) and a target age \( Age_{\text{target}} \), we compute a weighting factor based on the relative position between the input age \( Age_{\text{input}} \), the target age \( Age_{\text{target}} \), the upper reference bound \( Age_{\text{high}} \) (e.g., 70), and the lower reference bound \( Age_{\text{low}} \) (e.g., 30), and apply this to modulate the inversion features:
\begin{align}
w &= \frac{Age_{\text{target}} - Age_{\text{input}}}{Age_{\text{high}} - Age_{\text{low}}} \label{eq:age_weight} \\
V_{\text{inv}} \leftarrow V_{\text{inv}} + & w \cdot \Delta V_{\text{ref}}, \quad
K_{\text{inv}} \leftarrow K_{\text{inv}} + w \cdot \Delta K_{\text{ref}} \label{eq:feature_update}
\end{align}

This process effectively aligns the attention features with expected semantic changes due to aging. For instance, transforming a 30-year-old face to appear 50 years old would involve applying half the computed age direction: $V_{\text{inv}} = V_{\text{inv}} + 0.5 \Delta V_{\text{ref}}$. This guidance leads to smoother, more realistic aging transformations, especially in challenging mid-life edits, while retaining the flexibility of text-driven conditioning and maintaining identity fidelity.

\section{Experiments}
\label{sec:experiments}

We first discuss the evaluation setup in Sec. \ref{ssec:setup}, including the Flux ODE solver we use, the methods we compare to, the conditions we evaluated on, and the metrics we use for numerical evaluation. In Sec. \ref{ssec:results_celeb} and Sec. \ref{ssec:result-nonceleb}, we show details of numerical and visual comparisons of our method with other state-of-the-art methods on celebrity and non-celebrity images respectively. In  Sec. \ref{ssec:userstudy}, we present a user study that shows how humans evaluate the performance of our method against other approaches. Finally, in Sec. \ref{ssec:ablation} we ablate the contributions of our attention mixing and regularization strategies.

\subsection{Experimental Setup}
\label{ssec:setup}

\textbf{Dataset.}
We evaluate our method on the same celebrity dataset used in the MyTM paper~\cite{qi_mytimemachine_2024}, which contains a curated set of real-life 12 celebrities spanning various age ranges. To further assess the robustness and generalization of our method, we additionally collect a set of 11 non-celebrity individuals with age annotations.

\noindent
\textbf{Baselines.}
Our method is built upon RF-Solver-Edit~\cite{wang_taming_2024}, employing its second-order RF-Solver for both inversion and denoising. We compare against several recent open-source flow-based image editing approaches: RF-Inversion~\cite{rout_semantic_2024}, RF-Solver-Edit~\cite{wang_taming_2024}, FlowEdit~\cite{kulikov_flowedit_2024}, FireFlow~\cite{deng2024fireflowfastinversionrectified}, and FlowChef~\cite{patel2024flowchef}. All baselines are evaluated using the same base model, Flux.1-dev, and under the same set of aging-related conditions: {\textit{alcoholism}, \textit{gain weight}, \textit{good skin care}, \textit{poor skin care}, \textit{hair loss}}, \textit{strong sunlight exposure}, and \textit{living in dry windy climate}. To ensure a fair comparison, we apply the same prompt refinement strategy described in Sec.~\ref{ssec:prompt} across all models.
In addition, we compare with the state-of-the-art for age transformation FADING~\cite{Chen_2023_BMVC} that uses Null-Text Inversion (NTI)\cite{mokady_null-text_2022} to perform text-guided re-aging from a single input image. 



\noindent
\textbf{Metrics.}
We follow the evaluation framework used in Personalize Anything~\cite{feng_personalize_anything}, using the \textit{CLIP-T} score~\cite{radford_learning_2021} to assess alignment between the edited image and the text prompt. To measure aging quality, use report \textit{Age Mean Absolute Error ($\text{Age}_{MAE}$)}, which compares the target age to the predicted age from FP-Age~\cite{lin_fp-age_2022}, and \textit{Identity Similarity ($\text{ID}_{sim}$)}, computed as the ArcFace~\cite{deng_arcface_2022} embedding similarity between the edited image and reference images of the same person. For the celebrity dataset, we use images of the subject at the target age as references. For the non-celebrity dataset, where multi-age references are unavailable, we use the input image itself as the reference. In addition to quantitative metrics, we conduct a human evaluation comparing our method against three main baselines, RF-Solver-Edit~\cite{wang_taming_2024}, FlowEdit~\cite{kulikov_flowedit_2024}, and FADING~\cite{Chen_2023_BMVC}, to assess editing quality in terms of age accuracy, identity preservation, condition alignment, and overall preference.


\subsection{Evaluation on Celebrity Dataset}
\label{ssec:results_celeb}

\begin{figure*}[t]
    \centering    \includegraphics[width=\linewidth]{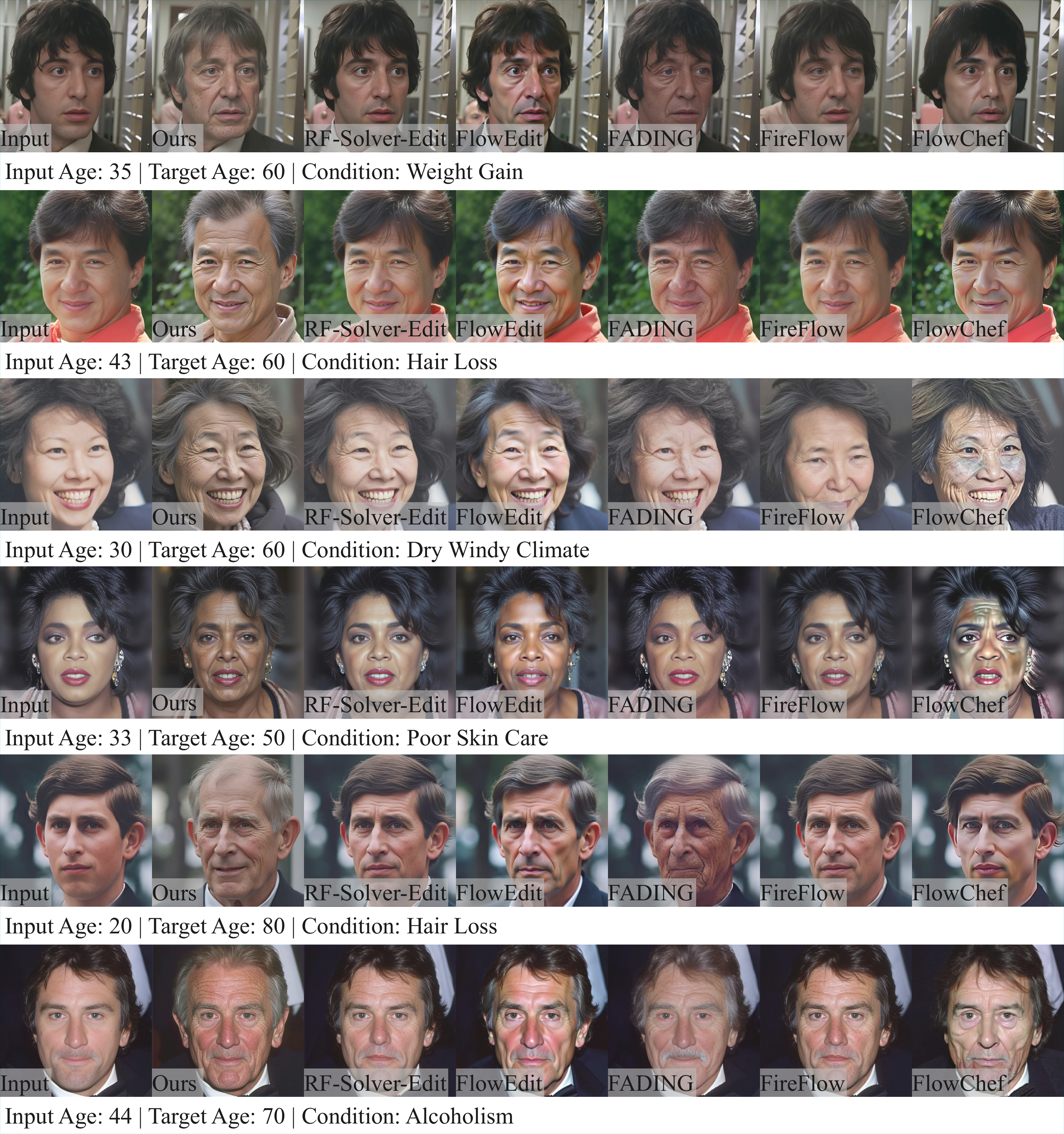}
\caption{
Given the input celebrity image on the left and the editing context indicated below each row, we present a visual comparison of our method with RF-Solver-Edit~\cite{wang_taming_2024}, FireFlow~\cite{deng2024fireflowfastinversionrectified}, FlowEdit~\cite{kulikov_flowedit_2024}, FlowChef~\cite{patel2024flowchef}, and FADING~\cite{Chen_2023_BMVC}. For FADING, only the aging effect is evaluated.
}
    \label{fig:visual_results_celeb}
\end{figure*}

We present a comparison of our method with recent flow-based image editing models and state-of-the-art age transformation methods, summarized in Fig.~\ref{fig:visual_results_celeb}. We evaluate performance on three key metrics: \textit{CLIP-T} for prompt alignment, \textit{Age MAE} for aging accuracy, and \textit{ID Similarity} for identity preservation in Table \ref{tab:celeb}.

Among the flow-based editing baselines, our method achieves the best overall balance between editability and identity preservation. It ranks highest in \textit{CLIP-T} (0.326), indicating strong alignment with the target prompt, and highest in \textit{Age MAE} (9.5), demonstrating accurate age transformation. While its \textit{ID Similarity} score (0.49) is lower than that of RF-Solver-Edit (0.57) and FireFlow (0.51), this is expected: those methods prioritize identity preservation by directly replacing attention features, as discussed in Sec.~\ref{ssec:att-mix}. 
As a result, they achieve near-perfect fidelity to the input image but show minimal responsiveness to editing prompts, as reflected in their significantly lower \textit{CLIP-T} scores and higher \textit{Age MAE} values.
Visual comparisons in Fig. \ref{fig:rf_comparison} and \ref{fig:visual_results_celeb} further illustrate this trade-off: models with high $\text{ID}_{\text{sim}}$ scores tend to overfit to the input appearance, producing outputs with minimal visual change. 
Our method, by contrast, produces edits that are both prompt-aligned and identity-aware.

Compared to FADING—an age transformation specialist—our outputs are visually comparable in terms of aging realism, even though FADING achieves slightly better \textit{Age MAE} and \textit{ID Similarity}. Importantly, FADING does not support conditional prompts (e.g., “60 years old with hair loss”), limiting its flexibility. Our approach not only edits toward target ages but also adapts to lifestyle or health-related conditions.

\begin{figure*}[t]
    \centering    \includegraphics[width=\linewidth]{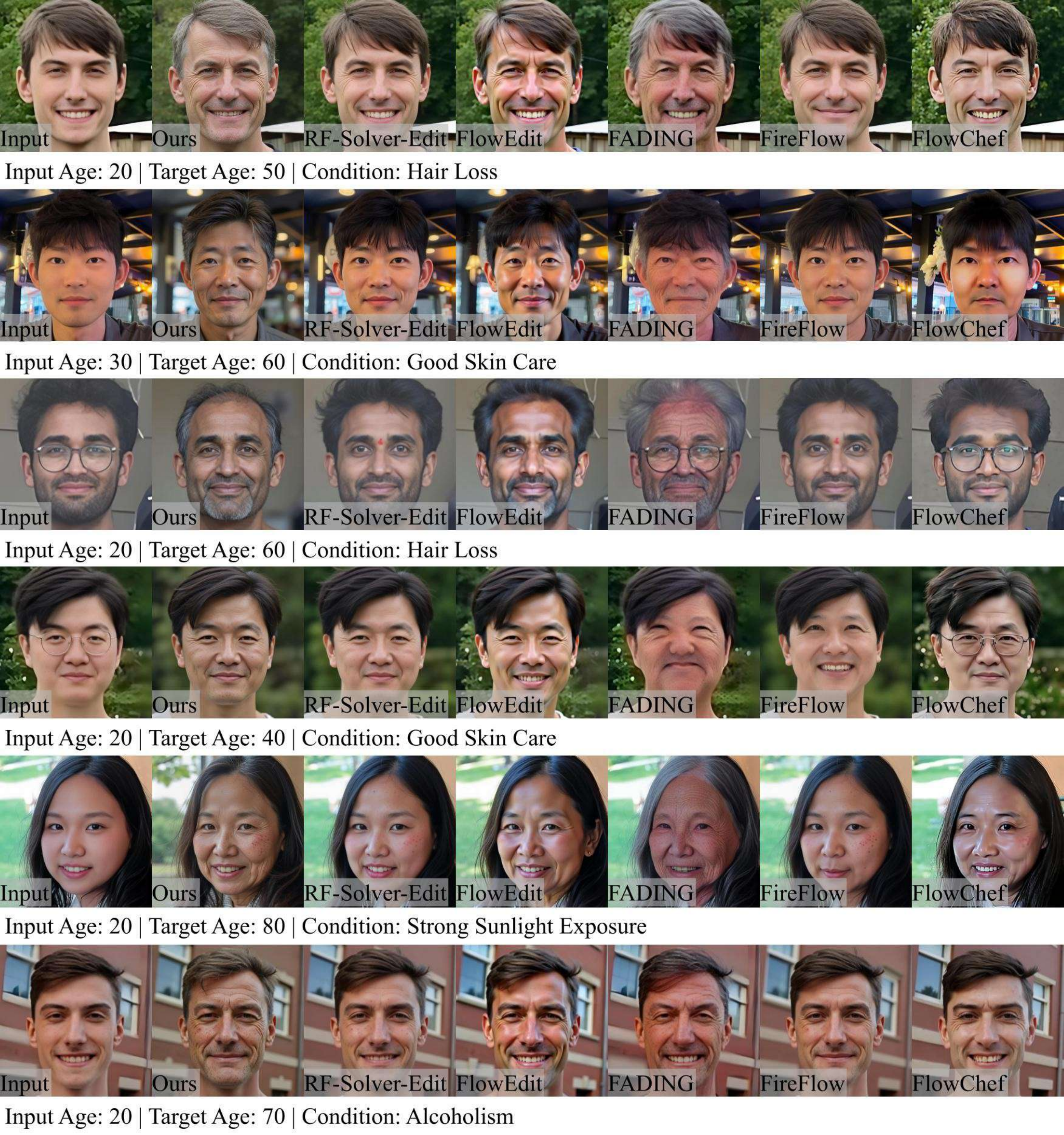}
    \caption{
Given the input in-the-wild non-celebrity image on the left and the editing context indicated below each row, we present a visual comparison of our method with RF-Solver-Edit~\cite{wang_taming_2024}, FireFlow~\cite{deng2024fireflowfastinversionrectified}, FlowEdit~\cite{kulikov_flowedit_2024}, FlowChef~\cite{patel2024flowchef}, and FADING~\cite{Chen_2023_BMVC}. For FADING, only the aging effect is evaluated.}
    \label{fig:visual_results_nonceleb}
\end{figure*}

\subsection{Evaluation on Non-celebrity Dataset}
\label{ssec:result-nonceleb}

\begin{table}[t]
    \centering
    \resizebox{\columnwidth}{!}{%
\begin{tabular}{lccc}
    \hline
    Method & CLIP-T ($\uparrow$) & Age$_{MAE}$ ($\downarrow$) & ID$_{sim}$ ($\uparrow$) \\
    \hline
    FADING* & - & \cellcolor{tabfirst}10.4 & \cellcolor{tabsecond}0.65 \\
    \hline
    RF-Inversion & 0.311 & 18.5 & 0.28 \\        
    FlowChef & \cellcolor{tabthird}0.312 & 18.1 & 0.47 \\
    FlowEdit & \cellcolor{tabsecond}0.322 & \cellcolor{tabthird}17.6 & 0.40 \\
    Fireflow & 0.300 & 23.3 & \cellcolor{tabthird}0.62 \\
    RF-Solver-Edit & 0.287 & 25.2 & \cellcolor{tabfirst}0.75 \\
    \hline
    Ours & \cellcolor{tabfirst}0.331 & \cellcolor{tabsecond}13.9 & 0.49 \\
    \hline
\end{tabular}
}

    \caption{Quantitative comparison of condition-aware age transformation on non-celebrity dataset. * indicates methods that require aging-specific pre-training and cannot be conditioned on external prompts; thus, no CLIP-T score is reported for these methods.
    \colorbox{tabfirst}{\raisebox{0pt}[1ex][0ex]{Red}} highlights the best result, 
\colorbox{tabsecond}{\raisebox{0pt}[1ex][0ex]{Orange}}  indicates the second best, 
and \colorbox{tabthird}{\raisebox{0pt}[1ex][0ex]{Yellow}} denotes the third best.}
    \label{tab:non_celeb}
    \vspace{-2em}
\end{table}

Similar to results in Sec.~\ref{ssec:results_celeb}, our method achieves the best balance across all three evaluation metrics for in-the-wild non-celebrities as shown in Table~\ref{tab:non_celeb}. It achieves the highest \textit{CLIP-T} score (0.331), indicating strong alignment with the target prompt, and ranks second in \textit{Age MAE} (13.9), behind FADING, which trains a dedicated aging model. While its \textit{ID Similarity} score (0.49) is not the highest, it still outperforms most flow-based baselines, which often suffer from poor identity preservation or limited editability.

For instance, RF-Solver-Edit and FireFlow achieve the highest identity scores (0.75 and 0.62, respectively), but their performance on \textit{CLIP-T} and \textit{Age MAE} indicates minimal responsiveness to prompts and weak aging accuracy, with visual examples shown in Fig.~\ref{fig:visual_results_nonceleb}. On the other hand, FlowEdit shows strong prompt alignment and age prediction but fails to preserve identity. Our method sits at a favorable point in this trade-off space, successfully editing facial attributes according to the prompt while maintaining visual consistency with the input identity. Compared to FADING, which benefits from extensive age-specific pretraining, our method achieves competitive aging performance without requiring any task-specific finetuning, underscoring its flexibility and robustness in a general editing framework.

\begin{figure}[t]
    \centering
    \includegraphics[width=0.9\linewidth]{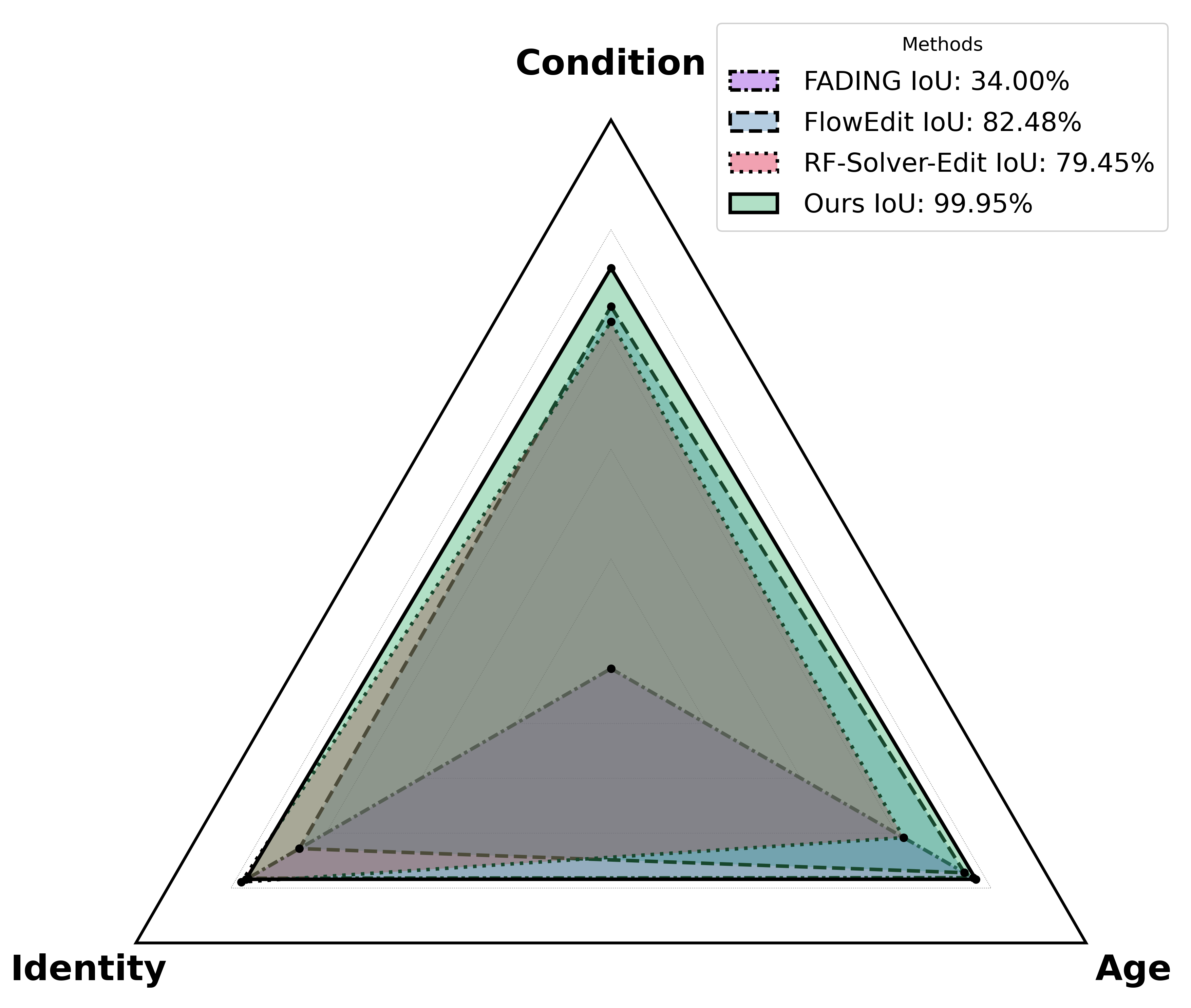}
    \vspace{-1em}
    \caption{Visualization of human evaluation scores across three axes: external condition alignment, age accuracy, and ID preservation from Table.~\ref{tab:celeb}. For each method, we calculate IoU as the ratio of area covered by the method's triangle over the union of all four triangles. This indicates our approach provides the best balance across all three criterion.
    }
    \label{fig:radar-plot}
    \vspace{-1em}
\end{figure}

\begin{figure}[t]
    \centering
    \includegraphics[width=\linewidth]{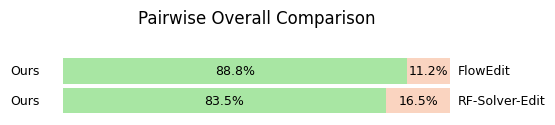}
    \vspace{-2em}
    \caption{
    Pairwise user study comparing our method to FlowEdit and RF-Solver-Edit to judge overall condition-aware age editing performance. 
    Values indicate the percentage of users preferring our method over each baseline.
    }
    \label{fig:user_study}
    \vspace{-1.5em}
\end{figure}

\subsection{User Study}
\label{ssec:userstudy}
Given the novelty of our task, we acknowledge that existing evaluation metrics are not always well-suited to capture the perceptual quality of edits. In particular, the \textit{ID Similarity} metric used in~\cite{qi_mytimemachine_2024}—based on ArcFace~\cite{deng_arcface_2022} embedding similarity between the edited image and either a reference or the input image—can introduce bias against our method. For celebrity data, the reference image corresponds to the target age, while for non-celebrities, the input image is used directly. This setup penalizes edits that introduce realistic aging and external condition effects not present in the real reference.
For example, in Fig.~\ref{fig:rf_comparison}, editing a young image of \textit{Al Pacino} to appear 60 years old with hair loss yields a lower similarity score when compared to actual photos of him at that age, since he did not exhibit hair loss in reality. Such discrepancies become even more pronounced under conditions like poor skin care or chronic sun exposure, which can drastically alter facial appearance in ways that diverge from available reference images.

To address these limitations and better assess perceptual quality, we conducted a user study aimed at capturing human judgments in evaluating the quality of condition-aware aging. We conducted a user study via Amazon Mechanical Turk to evaluate three key aspects of the edited images: alignment with external conditions, age accuracy, and identity preservation. Participants rated each image on a 1–5 scale across these criteria. We select 10 images of varying identity spanning 7 different external conditions for this user study. Each image was rated by 15 different users. We ask users to separately judge each of the three criterion: alignment with external conditions, age accuracy, and identity preservation in different HITs. Based on numerical results in Table~\ref{tab:celeb}, we identify FADING~\cite{Chen_2023_BMVC}, FlowEdit~\cite{kulikov_flowedit_2024} and RF-Solver-Edit~\cite{wang_taming_2024} as our main competitors.


The average scores are presented in Table~\ref{tab:celeb}, alongside the corresponding quantitative metrics.
Our method achieved the highest human ratings (Human Eval.) for both condition alignment and age accuracy, even slightly outperforming the aging-specialized model FADING. In terms of identity preservation, our score was slightly below that of RF-Solver-Edit, but notably closer than suggested by the embedding-based \textit{ID Similarity} metric. This highlights a key observation: our edits are perceived by humans as identity-consistent, despite being penalized by reference-based ID similarity metrics.
To visualize overall performance, we use a radar plot (Fig.~\ref{fig:radar-plot}) with three axes corresponding to the mean human scores on our evaluation criteria. Each method forms a triangle, and we compute the Intersection over Union (IoU) as the ratio of its area to the union of all method areas. Our method achieves the largest triangle with a 99.95\% IoU, reflecting balanced performance across all dimensions.

While per-criterion user ratings are informative, they don’t capture overall preferences—e.g., whether users favor minimal edits that preserve identity or more aggressive edits that compromise it. To assess holistic quality, we conducted a pairwise preference study comparing our method to FlowEdit and RF-Solver-Edit (excluding FADING, which cannot edit external conditions). Across 10 celebrity and 5 non-celebrity images, each rated by 10 users, participants preferred our results in 85\% of cases (Fig.\ref{fig:user_study}), highlighting our method’s strong balance between editability and identity fidelity.

\begin{figure}[t]
    \centering    
    \includegraphics[width=0.85\columnwidth]{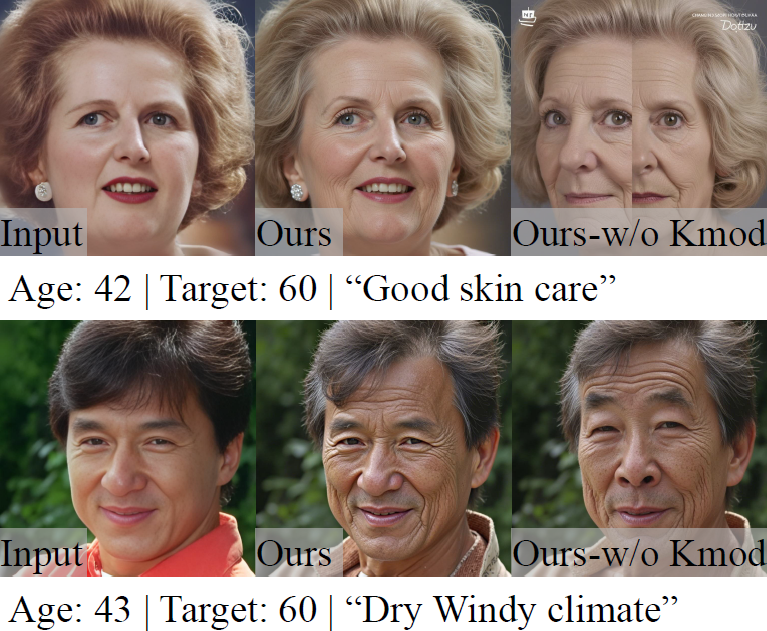}
    \vspace{-0.5em}
    \caption{
    Visual ablation of Attention Mixing with and without K modulation. Without K modulation, the edits are not stable, often leading to distorted face or our-of-distribution image.
    }
    \label{fig:k_motivation}
    \vspace{-0.5em}
\end{figure}

\begin{figure}[t]
    \centering    
    \includegraphics[width=0.85\columnwidth]{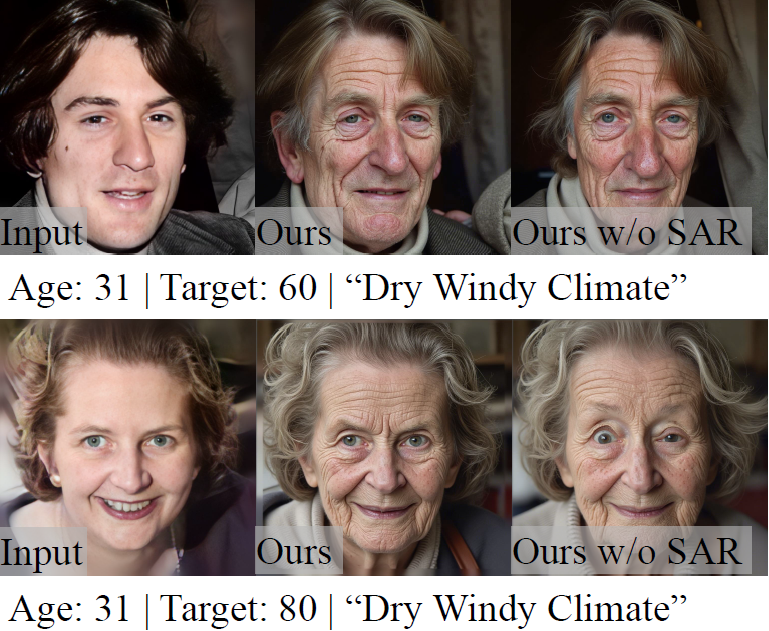}
    \vspace{-0.5em}
    \caption{
    Visual comparison between Ours and Ours w/o SAR. The results show that our proposed aging regulation helps stabilize editing results, preventing identity drift from the input image or unrealistic face distortions.
    }
    \label{fig:sar_figure}
    \vspace{-0.5em}
\end{figure}

\vspace{-0.5em}
\subsection{Ablation Study}
\label{ssec:ablation}
We perform a series of ablation studies to justify the effectiveness of our proposed components. Table~\ref{tab:ablation} presents the quantitative results, with each row incrementally building upon the RF-Solver-Edit~\cite{wang_taming_2024} baseline.
We begin by replacing the original value tensor copying strategy between identity and editing blocks with our proposed value projection approach, without masking out text embedding channels during the computation of $\alpha$. This change improves both \textit{CLIP-T} and \textit{Age$_{\text{MAE}}$}, indicating stronger prompt alignment and more accurate aging. As expected, \textit{ID$_\text{sim}$} decreases due to the increased influence of semantic edits, consistent with our earlier observations.

Next, we introduce the text embedding masking strategy (Sec.~\ref{sec:method}) to disentangle identity and prompt information. This further boosts \textit{CLIP-T} to 0.317 and reduces \textit{Age${\text{MAE}}$} to 12.5, but again slightly lowers \textit{ID Similarity}, as the edits become more visually distinct. However, as illustrated in Fig.~\ref{fig:k_motivation}, this increase in editing power may occasionally lead to instability or unrealistic outputs.
To mitigate these issues, we first incorporate the Key modulation technique to further improve the editing quality as shown in Fig. \ref{fig:k_motivation}. Finally, our Simulated Aging Regularization (SAR) further enhances edit stability while partially improving identity preservation, as shown in Fig.~\ref{fig:sar_figure}. This final version of our method achieves strong semantic edits with improved robustness, validating the effectiveness of our design choices.

\begin{table}[t]
    \centering
    \resizebox{\columnwidth}{!}{%
    \begin{tabular}{lcccc}
        \hline
        Method & CLIP-T$\uparrow$ & Age$_{MAE}$$\downarrow$ & ID$_{sim}$$\uparrow$ \\
        \hline
        RF-Solver-Edit (baseline) & 0.292 & 17.8 & 0.57 \\
         + Att. Mixing (Value only) & 0.304 & 14.4 & 0.50 \\
         + Text Embedding Masking & 0.317 & 12.5 & 0.47 \\
         + Att. Mixing (Value \& Key) & 0.322 & 11.0 & 0.48 \\
         + Simulated Aging Regularization & 0.326 & 9.5 & 0.49 \\
        \hline
    \end{tabular}
    }
    \caption{
    Ablation study validating the effectiveness of attention mixing (dynamic feature modulation), text channel masking, and attention regularization (simulated aging regularization).
    }
    \label{tab:ablation}
    \vspace{-1em}
\end{table}

\begin{figure}[t]
    \centering    
    \includegraphics[width=\columnwidth]{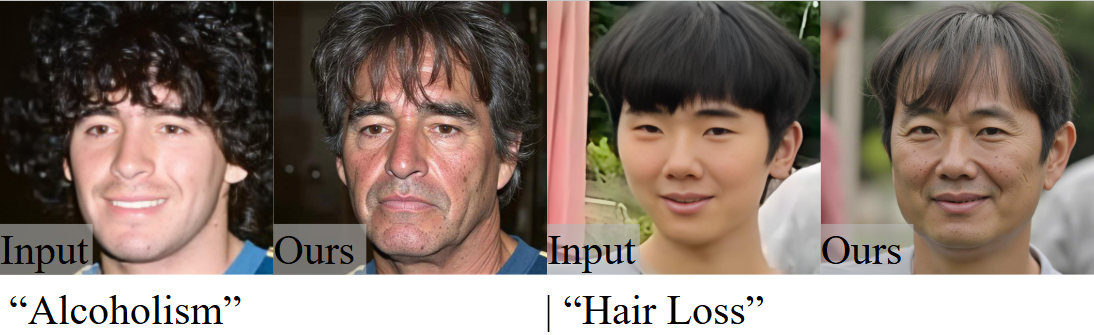}
    \vspace{-1.5em}
    \caption{
    Failure cases of our method when input image is low-quality, which often leads to identity drift from input image or desired edits not generated.
    }
    \label{fig:failure}
    \vspace{-1.5em}
\end{figure}

\section{Conclusion}
\label{sec:conclusion}

In this paper, we define the novel task of diverse-conditioned age transformation and present a simple, training-free method that extends flow-based models like Flux to this problem. By manipulating attention during inference, our approach adds minimal computational overhead while effectively combining identity preservation with text-driven edits. It enables the generation of aging trees that reflect various lifestyle and environmental conditions. Our method broadens the scope of facial aging research, achieving strong performance across both qualitative and quantitative benchmarks, outperforming prior Flux-based editors and matching the capabilities of state-of-the-art aging models.

\noindent \textbf{Limitation.} Although we introduced multiple strategies to stabilize the attention mixing, our method is still input-sensitive. As shown in Fig.~\ref{fig:failure}, our method could fail on low-quality heavily pre-processed images by changing the identity or failing to generate desired edit, especially when the edit requires significant change.

\noindent \textbf{Ethical Consideration.} The photorealistic facial aging trajectories generated by our method are algorithmic simulations, not deterministic predictions. While the trajectory is plausible and can be used for visual effects applications, it should not be used for facial identification purposes as it may likely raise many false positives. Rigorous human subject testing must be performed before such a tool can be deployed for health education or lifestyle choice determination.

\section*{Acknowledgments}
This research was supported in part by Lenovo Research (Morrisville, NC). We gratefully acknowledge the invaluable support and assistance of the members of the Mobile Technology Innovations Lab. This work was also supported in part by the National Science Foundation under Grant No. 2213335.




\bibliographystyle{ieeenat_fullname}
\bibliography{references, references_new, sample-base}

\appendix

\end{document}